\pdfoutput=1
\documentclass[10pt, logo, copyright]{nv}

\usepackage{wrapfig}
\usepackage{pgfplots}
\usepackage{algorithm}
\usepackage{algorithmic}
\usepackage{subcaption}
\usepackage{natbib}

\usepgfplotslibrary{groupplots}
\pgfplotsset{compat=1.18}

\newcommand{\modelname}{Fast-dVLM}
\definecolor{bestbg}{RGB}{220,235,255}
\definecolor{sndbg}{RGB}{240,245,255}
\newcommand{\bestd}[1]{\cellcolor{bestbg}\textbf{#1}}
\newcommand{\sndd}[1]{\cellcolor{sndbg}\underline{#1}}

\definecolor{darkblue}{rgb}{0, 0, 0.5}

\title{\modelname{}: Efficient Block-Diffusion VLM via Direct Conversion from Autoregressive VLM}

\author{
Chengyue Wu\textsuperscript{1,2*} ~~
Shiyi Lan\textsuperscript{2*} ~~
Yonggan Fu\textsuperscript{2} ~~
Sensen Gao\textsuperscript{4} ~~
Jin Wang\textsuperscript{1,2} ~~
Jincheng Yu\textsuperscript{2} ~~
Jose M. Alvarez\textsuperscript{2} ~~
Pavlo Molchanov\textsuperscript{2} ~~
Ping Luo\textsuperscript{1} ~~
Song Han\textsuperscript{2,3} ~~
Ligeng Zhu\textsuperscript{2\dag} ~~
Enze Xie\textsuperscript{2\dag} \\
\vspace{2mm}
{\normalsize
\textsuperscript{1}The University of Hong Kong ~~
\textsuperscript{2}NVIDIA ~~
\textsuperscript{3}MIT ~~
\textsuperscript{4}MBZUAI \\
\textsuperscript{*}Equal contribution \quad \textsuperscript{\dag}Co-lead
}
}

\begin{abstract}
\textbf{Abstract:} Vision-language models (VLMs) predominantly rely on autoregressive decoding, which generates tokens one at a time and fundamentally limits inference throughput.
This limitation is especially acute in physical AI scenarios such as robotics and autonomous driving, where VLMs are deployed on edge devices at batch size one, making AR decoding memory-bandwidth-bound and leaving hardware parallelism underutilized.
While block-wise discrete diffusion has shown promise for parallel text generation, extending it to VLMs remains challenging due to the need to jointly handle continuous visual representations and discrete text tokens while preserving pretrained multimodal capabilities.
We present \textbf{\modelname{}}, a block-diffusion-based VLM that enables KV-cache-compatible parallel decoding and speculative block decoding for inference acceleration.
We systematically compare two AR-to-diffusion conversion strategies: a \emph{two-stage} approach that first adapts the LLM backbone with text-only diffusion fine-tuning before multimodal training, and a \emph{direct} approach that converts the full AR VLM in one stage. Under comparable training budgets, direct conversion proves substantially more efficient by leveraging the already multimodally aligned VLM; we therefore adopt it as our recommended recipe.
We introduce a suite of multimodal diffusion adaptations---block-size annealing, causal context attention, auto-truncation masking, and vision-efficient concatenation---that collectively enable effective block diffusion in the VLM setting.
Extensive experiments across 11 multimodal benchmarks show \modelname{} matches its autoregressive counterpart in generation quality. With SGLang integration and FP8 quantization, \modelname{} achieves over $6\times$ end-to-end inference speedup over the AR baseline.
\newline
\textbf{Links:} \hspace{2pt}
{\hypersetup{urlcolor=nvidiagreen}
\href{https://github.com/NVlabs/Fast-dLLM}{Github Code} |
\href{https://nvlabs.github.io/Fast-dLLM/fast_dvlm} {Project Page}
}
\end{abstract}

\begin{document}

\maketitle

\definecolor{refblue}{RGB}{95,140,200}
\definecolor{reforange}{RGB}{230,145,56}
\definecolor{refgreen}{RGB}{145,190,70}
\definecolor{refred}{RGB}{205,100,100}
\definecolor{refpurple}{RGB}{160,110,160}
\definecolor{refbrown}{RGB}{165,130,110}

\begin{figure*}[h!]
  \centering
  \includegraphics[width=\textwidth]{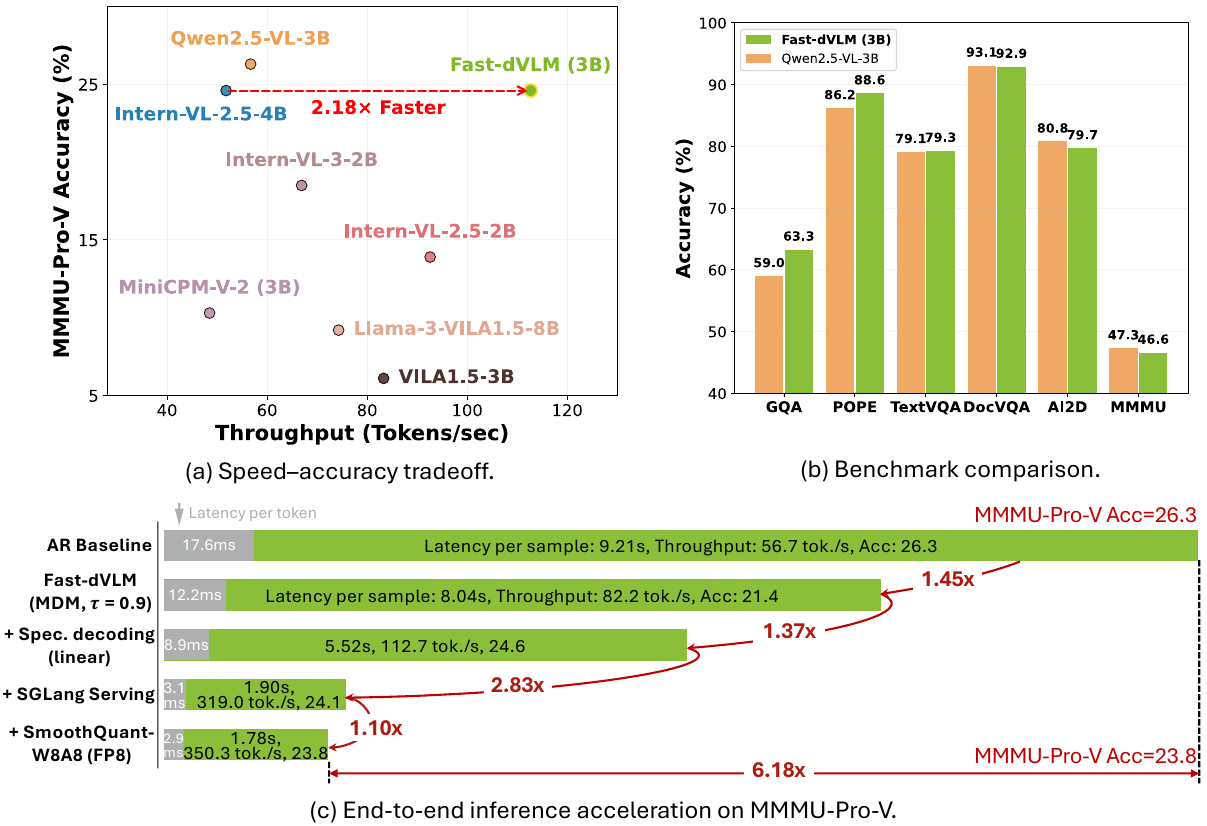}
  \caption{
    Overview of our \modelname{}.
    (a) \modelname{} achieves comparable accuracy to AR VLM baselines on MMMU-Pro-V with a substantial speedup. (b) Benchmark comparison against the Qwen2.5-VL-3B backbone, showing near-lossless accuracy across diverse tasks. (c) Our combined approach achieves up to 6.18$\times$ end-to-end speedup compared to the AR baseline. All throughput measurements are conducted on a single NVIDIA H100 GPU.
  }
  \label{fig:teaser}
  \vspace{-15pt}
\end{figure*}

\section{Introduction}

Vision-language models (VLMs)~\citep{grattafiori2024llama,wang2024qwen2,bai2025qwen25vl} have become the foundation for a wide
range of multimodal applications, from visual question answering to document
understanding and chart interpretation.
As these models are deployed in increasingly demanding settings, generating
long-form reasoning chains, structured outputs, and multi-turn dialogues,
inference efficiency becomes a critical bottleneck.
Current VLMs almost universally rely on autoregressive (AR) decoding, producing
tokens one at a time, which fundamentally limits generation throughput regardless
of available hardware parallelism.

This efficiency gap is especially critical in physical AI, an increasingly important deployment frontier for VLMs.
In robotics, autonomous driving, and embodied agents, VLMs serve as core perception-reasoning modules that must operate on resource-constrained edge platforms.
Unlike cloud-serving workloads that amortize the cost of sequential decoding across large request batches, physical AI deployments are dominated by single-request, batch-size-one inference: each robot or vehicle processes its own observation stream independently.
In this regime, AR decoding is fundamentally \emph{memory-bandwidth-bound}---the model must load its full parameters for every generated token, yet utilizes only a small fraction of the available compute.
Diffusion-based parallel decoding offers a natural remedy: by generating multiple tokens simultaneously within each block, it shifts the workload toward a more compute-bound regime and can therefore better exploit hardware parallelism, even under tight batch-size-one constraints.

Diffusion-based language models~\citep{sahoo2024simple,nie2025large,ye2025dream,arriola2025block} have emerged as a promising alternative, enabling parallel multi-token decoding, with speculative decoding~\citep{gao2025self,agrawal2025spiffy,pan2025blockspec,chen2026dflash,pan2025failfast} providing further speedups.
However, these advances remain largely confined to text-only settings.

Extending diffusion generation to VLMs~\citep{you2025llada,li2025lavida,yang2025mmada,yu2025dimple,zeng2025diffusionvl,arriola2025ar2d,cheng2025sdarvl} is challenging, as the model must jointly handle visual and textual modalities while preserving pretrained multimodal capabilities.
Early efforts~\citep{you2025llada,li2025lavida,yang2025mmada,yu2025dimple} adopt full-sequence diffusion without block structure, precluding incremental KV caching; more recent works~\citep{zeng2025diffusionvl,arriola2025ar2d,cheng2025sdarvl} introduce block diffusion with KV cache reuse.
However, a key question remains underexplored: is it more effective to first adapt an AR LLM into a diffusion LLM and then fine-tune multimodally, or to directly convert a pretrained AR VLM in a single step?

In this work, we present \textbf{\modelname{}}, a block-diffusion-based
vision-language model that, beyond KV-cache-compatible parallel decoding, further
contributes self-speculative
block decoding and system-level integration with SGLang~\citep{zheng2024sglang} for production-grade inference acceleration.
At the core of \modelname{} is a systematic comparison of two AR-to-diffusion conversion strategies:
a \emph{two-stage} path that first converts the LLM backbone via text-only diffusion fine-tuning before multimodal fine-tuning, and a \emph{direct} path that converts the full VLM in a single multimodal stage.

Our comparison reveals that, under comparable training budgets, direct conversion is significantly more training-efficient: by starting from a multimodally aligned VLM rather than a text-only LLM, it makes better use of the same data and compute. We hypothesize that both strategies share a similar performance ceiling but differ primarily in how efficiently they utilize the training budget, and adopt the direct path as our recommended recipe.
Both paths build on Fast-dLLM v2~\citep{wu2025fastv2} and extend it to VLMs with causal context attention, auto-truncation masking, vision-efficient concatenation, and self-speculative decoding for additional inference speedup.

Our contributions are threefold:
\begin{itemize}
  \item We propose \textbf{\modelname{}}, a block-diffusion VLM compatible with KV caching and self-speculative decoding that achieves competitive quality with significant inference speedups over AR baselines.
  \item Through a controlled comparison of two-stage and direct AR-to-diffusion conversion, we find the direct path both simpler and more effective, and propose a systematic recipe---block-size annealing, auto-truncation masking, vision-efficient concatenation, and self-speculative decoding---each validated by ablation.
  \item Extensive experiments across 11 multimodal benchmarks show that \modelname{} matches or surpasses its AR counterpart. Combined with SGLang~\citep{zheng2024sglang} integration and FP8 quantization, \modelname{} achieves up to $6\times$ end-to-end inference speedup.
\end{itemize}

\section{Related Work}

Discrete diffusion language models have progressed from continuous-latent formulations~\citep{lovelace2023latent,strudel2022self} to masked diffusion objectives~\citep{austin2021structured,lou2023discrete,sahoo2024simple}, with LLaDA~\citep{nie2025large} and Dream~\citep{ye2025dream} scaling to 7--8B parameters and matching AR baselines.
Block-level generation~\citep{arriola2025block,wu2025fast,wu2025fastv2,gat2025sbd} further enables KV caching by autoregressively producing blocks whose tokens are denoised in parallel.
Several concurrent works extend diffusion to VLMs---LLaDA-V~\citep{you2025llada}, LaViDa~\citep{li2025lavida}, MMaDA~\citep{yang2025mmada}, Dimple~\citep{yu2025dimple}---but all rely on full-sequence diffusion without block structure, precluding incremental KV caching and turn-aware attention masking.
Speculative decoding for dLLMs~\citep{gao2025self,agrawal2025spiffy,pan2025blockspec,chen2026dflash,pan2025failfast,li2025diffuspec} exploits native multi-token prediction for further speedup.
Our work is the first to integrate block-level self-speculation into a multimodal diffusion VLM with system-level serving support.
A detailed version is provided in Appendix~\ref{app:related_work}.

\section{Methodology}

\subsection{Preliminary}
\label{sec:prelim_dvlm_extended}

Let $\mathbf{x}=(x_1,\dots,x_L)$ denote a token sequence of length $L$.
Autoregressive models generate $\mathbf{x}$ by factoring $P_\theta(\mathbf{x})=\prod_i P_\theta(x_i\mid x_{<i})$.
Masked diffusion models~\citep{sahoo2024simple,nie2025large} instead corrupt each token independently with masking probability $t\in(0,1)$ and train a reverse model $p_\theta(\mathbf{x}_0\mid\mathbf{x}_t)$ to recover masked tokens.
Block-wise discrete diffusion~\citep{arriola2025block,wu2025fastv2} partitions the sequence into blocks of size $\hat{B}$, generating blocks autoregressively while denoising tokens in parallel within each block, enabling KV-cache reuse across blocks.

Our model builds on Fast-dLLM v2~\citep{wu2025fastv2}, inheriting its complementary masking and dual-stream (noisy + clean) block attention design.
Extending this text-only framework to VLMs raises four challenges:
\textbf{(1)~Conversion strategy}---should a pretrained AR VLM be converted via a two-stage text-first path or directly in one step?
\textbf{(2)~Multi-turn boundaries}---short responses (e.g., a single letter) cause the last denoising block to extend into the next turn's prompt, leaking future information.
\textbf{(3)~Training efficiency}---the noisy-clean concatenation duplicates vision embeddings into both streams even though they are never corrupted, wasting memory and compute.
\textbf{(4)~Causal compatibility}---block-level context attention discards the pretrained causal structure and precludes AR decoding for self-speculative verification.
The following subsections address each challenge.

\begin{figure*}[t]
  \centering
  \includegraphics[width=0.9\columnwidth]{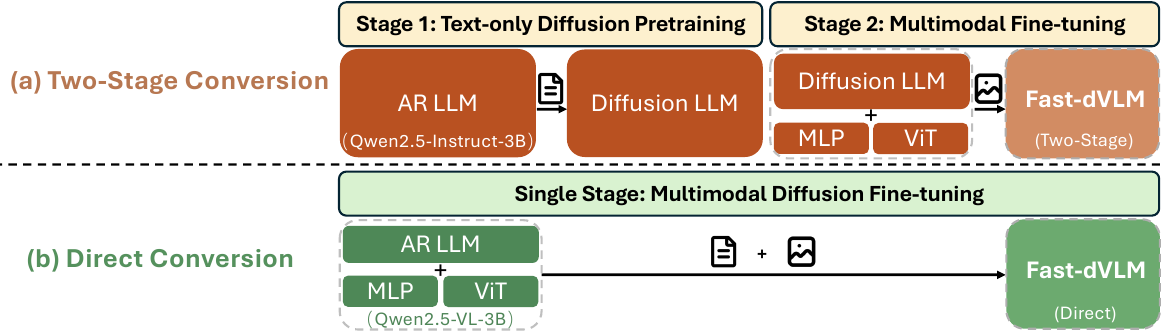}
  \caption{%
      Two AR-to-diffusion conversion strategies.
      (a)~The \emph{two-stage} strategy first converts the LLM backbone via
      text-only diffusion fine-tuning~(Stage~1), then fine-tunes all
      components on multimodal data~(Stage~2).
      (b)~The \emph{direct path} converts the full VLM in a single
      multimodal stage.%
    }
    \vspace{-10pt}
  \label{fig:dual_path}
\end{figure*}

\subsection{AR-to-Diffusion Conversion Strategies}
\label{sec:dual_path}

We compare two strategies for converting a pretrained AR VLM into a block-diffusion model (Figure~\ref{fig:dual_path}):

\paragraph{Two-stage path.}
Stage~1 converts only the LLM backbone via text-only diffusion fine-tuning; Stage~2 attaches the vision encoder and MLP projector and jointly fine-tunes the entire model on multimodal data, with the vision encoder unfrozen since the LLM has already stabilized under the diffusion objective.

\paragraph{Direct path.}
The complete AR VLM is directly fine-tuned for block-diffusion on multimodal data in a single stage, yielding a simpler pipeline that leverages the pretrained multimodal alignment.
Our experiments (Section~\ref{sec:experiments}) show that, under comparable training budgets, the direct path achieves substantially stronger performance, suggesting both strategies share a similar ceiling but the direct path makes more efficient use of its multimodal initialization.

\begin{figure}[t]
\centering
\includegraphics[width=0.85\textwidth]{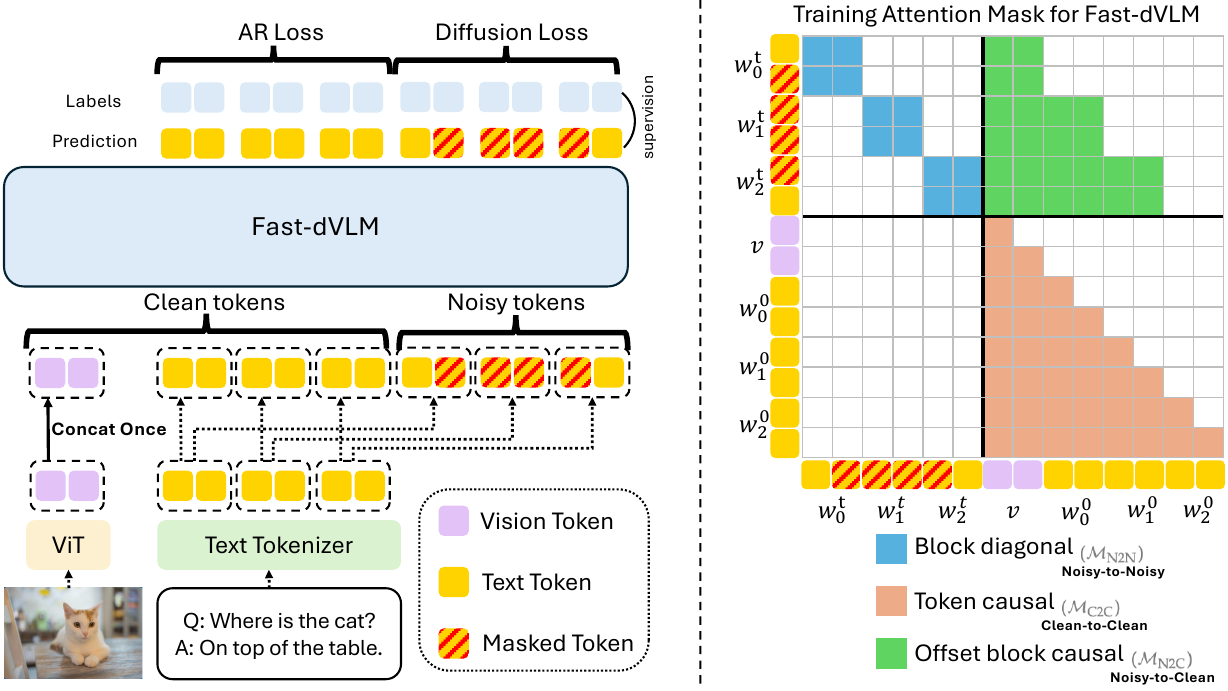}
\caption{%
  Training architecture and attention mask for $[\mathbf{w}^t;\mathbf{x}]$ with block size $\hat{B}=2$.
  The noisy stream $\mathbf{w}^t$ contains only text tokens; vision tokens appear exclusively in the clean stream $\mathbf{x}$ (vision-efficient concatenation).
  $\mathcal{M}_{\mathrm{N2N}}$: block-diagonal attention for parallel denoising.
  $\mathcal{M}_{\mathrm{N2C}}$: noisy tokens attend to preceding clean context including vision tokens.
  $\mathcal{M}_{\mathrm{C2C}}$: token-level causal attention on the clean stream, enabling joint AR loss training and AR decoding at inference.
}
\vspace{-10pt}
\label{fig:attention_mask}
\end{figure}

\subsection{Training}
\label{sec:training}

Both paths share the same training pipeline.
Let $\mathbf{x}=(\mathbf{v},\mathbf{w})$ denote the full input sequence, where $\mathbf{v}$ are vision token embeddings and $\mathbf{w}$ are text token embeddings.
Only response text tokens are corrupted: a noisy stream $\mathbf{w}^t$ is constructed by masking response positions and concatenated with the clean stream to form $[\mathbf{w}^t;\mathbf{x}]$.

The attention mask (Figure~\ref{fig:attention_mask}) enforces three rules:
noisy tokens attend bidirectionally within their block ($\mathcal{M}_{\mathrm{N2N}}$);
noisy tokens attend to clean tokens from preceding blocks ($\mathcal{M}_{\mathrm{N2C}}$);
clean tokens follow token-level causal attention ($\mathcal{M}_{\mathrm{C2C}}$).
Unlike Fast-dLLM v2's block-level context attention, we use causal attention for all preceding context, which better preserves pretrained AR representations and supports AR decoding for self-speculative verification.

\paragraph{Block-size annealing.}
\label{sec:annealing_block}
We adopt a curriculum that progressively increases the block size $\hat{B}$.
Given candidate sizes $S=\{2^1,2^2,\dots,B_d\}$ and training progress $u\in[0,1]$, the active size is $\hat{B}=S[\min(\lfloor u\cdot|S|\rfloor,\;|S|-1)]$, allowing the model to learn fine-grained denoising before tackling larger corruption spans.

\paragraph{Auto-truncation attention mask.}
\label{sec:masking}
In multi-turn dialogue, response lengths are rarely multiples of $\hat{B}$---especially in multimodal data where responses can be extremely short (e.g., a single option letter).
Without special handling, the last block would extend into the next turn's prompt, and $\mathcal{M}_{\mathrm{N2N}}$ would let noisy tokens attend to future prompt tokens.
We address this by automatically truncating each response's last block at the response boundary, preventing cross-turn leakage while preserving block-parallel denoising.

\paragraph{Vision-efficient concatenation.}
\label{sec:vision_preserving}
Since vision embeddings are never corrupted, they are identical across both streams.
We therefore include them only in the clean stream; the noisy stream contains only text positions, and vision tokens are attended to via $\mathcal{M}_{\mathrm{N2C}}$.
On Qwen2.5-VL-3B (H100, context length 2048), this lossless design reduces peak memory by 15.0\% and training time by 14.2\%.

\paragraph{Training objective.}
\label{sec:objective}
Let $W$ denote the language model head, and $\mathbf{H}^{(t)}$, $\mathbf{H}^{(0)}$ the hidden states of the noisy and clean streams.
The total loss combines a diffusion loss with a causal LM loss ($\alpha=\beta=0.5$):
\[
\mathcal{L} = \alpha\,\mathrm{CE}\!\bigl(W\mathbf{H}^{(t)},\,\mathbf{y}\bigr) +
\beta\,\mathrm{CE}\!\bigl(W\mathbf{H}^{(0)},\,\mathbf{y}\bigr),
\]
where $\mathbf{y}$ are the response labels.
The diffusion branch learns parallel denoising while the causal branch preserves AR generation capability.

\subsection{Inference}
\label{sec:inference_prelim}

As in Fast-dLLM v2~\citep{wu2025fastv2}, decoding proceeds block by block with KV-cache reuse; within each block, tokens are iteratively unmasked via confidence-aware parallel decoding (threshold $\tau$).
Our pipeline adds two key components: causal context decoding and self-speculative decoding.

\paragraph{Causal context decoding.}
Each block is seeded by a single AR step that generates the first token from the cached causal context; the remaining $\hat{B}-1$ positions are filled with \texttt{[MASK]} and iteratively denoised, naturally aligning with the training-time causal attention pattern.

\paragraph{Self-speculative block decoding.}
We introduce a self-speculative variant~\citep{samragh2025your,chen2026dflash} where the diffusion mode drafts all $\hat{B}-1$ tokens in one pass and the causal mode verifies them autoregressively, accepting the longest matching prefix and trimming the KV cache accordingly.
We employ a \emph{linear} scheme (two passes per block: draft + verify) and a \emph{quadratic} scheme~\citep{liu2025tidar} that fuses verification and next-block proposal into one pass with $O(\hat{B}^2)$ input tokens.
Details and pseudo code are in Appendix~\ref{sec:spec_details}.

\paragraph{System integration.}
We integrate \modelname{} into SGLang~\citep{zheng2024sglang}, extending its scheduler to support alternating bidirectional-draft and causal-verify attention, enabling optimized kernels and CUDA graph for wall-clock speedups in production serving. We further incorporate SmoothQuant~\citep{xiao2023smoothquant} to enable W8A8 (FP8) quantization, reducing memory footprint and improving effective Tensor Core throughput.

\begin{table*}[t] 
  \centering
  \caption{% 
    Benchmark performance comparison (Part~1: short-answer benchmarks). 
    Models are grouped into autoregressive (AR) and diffusion categories.
    Among diffusion models, \colorbox{bestbg}{\textbf{best}} and \colorbox{sndbg}{\underline{2nd best}} results are highlighted.
    MDM = masked diffusion model decoding; spec.\ = speculative decoding.%
  }
  \label{tab:main_results} 
  \vspace{4pt} 
  \resizebox{0.8\textwidth}{!}{% 
    \begin{tabular}{l *{7}{c}} 
      \toprule
      & \multicolumn{7}{c}{\textit{Short-answer Benchmarks}} \\ 
      \cmidrule(lr){2-8} 
      Model
        & AI2D & ChartQA & DocVQA & GQA & MMBench
        & MMMU & POPE \\ 
      \midrule
      \multicolumn{8}{c}{\textit{\textcolor{red!70!black}{Autoregressive Vision-Language Models}}} \\
      \midrule
      VILA-1.5-3B
        & 58.0 & 53.0 & 44.3 & 61.4 & 60.5
        & 31.8 & 86.8 \\ 
      MiniCPM-V-2 (3B)
        & 65.0 & 59.2 & 69.8 & 51.7 & 66.3
        & 37.9 & 86.5 \\ 
      Intern-VL-2.5-4B
        & 81.3 & 77.8 & 91.1 & 61.0 & 80.7
        & 50.0 & 89.3 \\ 
      Qwen2.5-VL-3B
        & 80.8 & 84.0 & 93.1 & 59.0 & 76.9
        & 47.3 & 86.2 \\ 
      \midrule
      \multicolumn{8}{c}{\textit{\textcolor{darkblue}{Diffusion Vision-Language Models}}} \\
      \midrule
      LaViDa
        & 70.0 & 59.0 & 64.6 & 55.5 & 70.5
        & 43.3 & 81.4 \\ 
      Dimple
        & 74.4 & 63.3 & 37.7 & 59.2 & \sndd{74.6}
        & 45.2 & \sndd{86.2} \\ 
      LLaDA-V 
        & \sndd{77.8} & \sndd{78.3} & \sndd{83.9} & 53.4 & \bestd{82.9}
        & \bestd{48.6} & 81.8 \\
      \modelname{} (MDM) 
        & \bestd{79.7} & 82.8 & 92.1
        & \sndd{63.0} & 74.2
        & 44.6 & \bestd{88.6} \\ 
      \modelname{} (spec.) 
        & \bestd{79.7} & \bestd{83.1} & \bestd{92.9}
        & \bestd{63.3} & 74.3
        & \sndd{46.6} & \bestd{88.6} \\ 
      \bottomrule
    \end{tabular}% 
  }% 
\end{table*} 

\begin{table*}[t] 
  \centering
  \caption{% 
    Benchmark performance comparison (Part~2).
    Tokens/NFE measures average tokens decoded per forward pass.
    Among diffusion models, \colorbox{bestbg}{\textbf{best}} and \colorbox{sndbg}{\underline{2nd best}} results are highlighted.
    MDM = masked diffusion model decoding; spec.\ = speculative decoding.%
  } 
  \label{tab:main_results_part2} 
  \vspace{4pt} 
  \resizebox{0.8\textwidth}{!}{% 
    \begin{tabular}{l *{4}{c} c c} 
      \toprule
      & \multicolumn{4}{c}{\textit{Short-answer Benchmarks}} 
      & \multicolumn{2}{c}{\textit{Long-answer}} \\ 
      \cmidrule(lr){2-5} \cmidrule(lr){6-7} 
      Model
        & RealWorldQA & SEEDBench2+ & TextVQA & Avg 
        & MMMU-Pro-V & Tokens/NFE \\ 
      \midrule
      \multicolumn{7}{c}{\textit{\textcolor{red!70!black}{Autoregressive Vision-Language Models}}} \\
      \midrule
      VILA-1.5-3B
        & 53.2
        & 41.2 & 58.2 & 54.8
        & 6.1 & 1.00 \\ 
      MiniCPM-V-2 (3B)
        & 56.3
        & 52.5 & 74.4 & 62.0
        & 10.3 & 1.00 \\ 
      Intern-VL-2.5-4B
        & 64.6
        & 67.0 & 78.8 & 74.2
        & 24.6 & 1.00 \\ 
      Qwen2.5-VL-3B
        & 65.1
        & 68.6 & 79.1 & 74.0
        & 26.3 & 1.00 \\ 
      \midrule
      \multicolumn{7}{c}{\textit{\textcolor{darkblue}{Diffusion Vision-Language Models}}} \\
      \midrule
      LaViDa
        & 54.5
        & 57.7 & 60.3 & 61.7
        & 10.5 & 1.00 \\ 
      Dimple
        & 55.4
        & 51.7 & 61.6 & 60.9
        & 12.4 & 1.00 \\ 
      LLaDA-V
        & \sndd{63.2}
        & \bestd{68.7} & 64.7 & 70.3
        & \sndd{18.6} & 1.00 \\ 
      \modelname{} (MDM)
        & \bestd{65.1} 
        & \sndd{67.2} & \sndd{76.1} & \sndd{73.3} 
        & 21.4 & \sndd{1.95} \\ 
      \modelname{} (spec.)
        & \bestd{65.1}
        & \sndd{67.2} & \bestd{79.3} 
        & \bestd{74.0} 
        & \bestd{24.6} & \bestd{2.63} \\ 
      \bottomrule
    \end{tabular}% 
  }% 
\end{table*}

\section{Experiments}
\label{sec:experiments}

\subsection{Experimental Setup}
\label{sec:exp_setup}

We initialize from Qwen2.5-VL-3B~\citep{bai2025qwen25vl} and convert it via the direct path (Section~\ref{sec:dual_path}) with all training recipes described in Section~\ref{sec:training} ($B_d{=}32$, $\alpha{=}\beta{=}0.5$).
At inference we evaluate MDM decoding and self-speculative decoding (linear variant by default; Algorithm~\ref{alg:linear-spec}).
We benchmark on 11 tasks all evaluated with VLMEvalKit~\citep{duan2024vlmevalkit}.
Throughput (TPS) is measured on a single NVIDIA H100 GPU at batch size one, reflecting the single-request regime prevalent in physical AI deployments where AR decoding is memory-bandwidth-bound.
We also report Tokens/NFE (average tokens decoded per forward pass; computed on MMMU-Pro-V samples with $>$200 response tokens).
Full experimental details are in Appendix~\ref{app:training_details}.

\subsection{Main Results}
\label{sec:main_results}

Table~\ref{tab:main_results} compares \modelname{} (trained with $B_d=32$) against AR and diffusion VLM baselines using two decoding strategies: MDM decoding and speculative (spec.) decoding.

\paragraph{Short-answer benchmarks.}
On short-answer tasks, \modelname{} achieves competitive results with the AR baseline while delivering significant inference speedup.
With MDM decoding, the model reaches an average score of 73.3, closely matching the baseline's 74.0, while achieving $1.95\times$ Tokens/NFE.
It matches or surpasses the baseline on GQA (+4.0), POPE (+2.4), and RealWorldQA, suggesting that bidirectional context within each denoising block benefits holistic visual reasoning.
With speculative decoding, the average score rises to 74.0, exactly matching the baseline, while Tokens/NFE reaches $2.63\times$.
Among diffusion VLMs, \modelname{} achieves the best results on 8 out of 11 short-answer benchmarks, substantially outperforming prior diffusion baselines.

\paragraph{Long-answer benchmarks.}
On MMMU-Pro-V, which requires multi-step chain-of-thought reasoning, the AR baseline scores 26.3.
\modelname{} with MDM decoding scores 21.4, a gap of 4.9 points; speculative decoding narrows this to 24.6 (only 1.7 points behind).
Long-form reasoning demands sequential coherence over many tokens, where block-wise parallel denoising is at a structural disadvantage.
The remaining gap can likely be further narrowed with larger-scale training data and longer annealing schedules.

\paragraph{Speed--quality tradeoff.}
A clear and favorable tradeoff emerges across both strategies:
MDM decoding provides $1.95\times$ Tokens/NFE with only a 0.7-point average quality drop on short-answer tasks, while speculative decoding achieves $2.63\times$ Tokens/NFE while exactly matching the baseline's average quality (74.0).
This confirms that block-wise discrete diffusion is a practical paradigm for VLMs, delivering acceleration without sacrificing benchmark accuracy on short-answer tasks.

\subsection{Direct vs.\ Two-Stage Adaptation}
\label{sec:path_comparison}

\begin{wrapfigure}{r}{0.5\columnwidth}
  \vspace{-12pt}
  \centering
\begin{tikzpicture}[scale=0.72, every node/.style={scale=0.72}]

\def\R{2.5}

\definecolor{colorA}{RGB}{31, 119, 180}
\definecolor{colorB}{RGB}{161, 32, 52}

\foreach \f in {0.25, 0.50, 0.75, 1.00} {
  \pgfmathsetmacro{\r}{\f * \R}
  \draw[gray!25, thin]
    (90:\r) -- (54:\r) -- (18:\r) -- ({-18}:\r) -- ({-54}:\r) --
    ({-90}:\r) -- ({-126}:\r) -- ({-162}:\r) -- (162:\r) -- (126:\r) -- cycle;
}

\foreach \i in {0,...,9} {
  \draw[gray!40, thin] (0,0) -- ({90 - \i * 36}:{\R});
}

\node[anchor=south,      font=\footnotesize] at ( 90:{\R+0.20}) {AI2D};
\node[anchor=south west, font=\footnotesize] at ( 54:{\R+0.15}) {ChartQA};
\node[anchor=west,       font=\footnotesize] at ( 18:{\R+0.15}) {DocVQA};
\node[anchor=west,       font=\footnotesize] at (-18:{\R+0.15}) {GQA};
\node[anchor=north west, font=\footnotesize] at (-54:{\R+0.15}) {MMBench};
\node[anchor=north,      font=\footnotesize] at (-90:{\R+0.20}) {MMMU};
\node[anchor=north east, font=\footnotesize] at ({-126}:{\R+0.15}) {POPE};
\node[anchor=east,       font=\footnotesize] at ({-162}:{\R+0.15}) {RealWorldQA};
\node[anchor=east,       font=\footnotesize] at (162:{\R+0.15}) {SeedBench2+};
\node[anchor=south east, font=\footnotesize] at (126:{\R+0.15}) {TextVQA};

\draw[colorA, semithick, fill=colorA, fill opacity=0.15]
  ({ 90}:{(79.7 - 55) / (86 - 55) * \R}) --
  ({ 54}:{(82.8 - 55) / (90 - 55) * \R}) --
  ({ 18}:{(92.1 - 50) / (103 - 50) * \R}) --
  ({-18}:{(63.0 - 50) / (66 - 50) * \R}) --
  ({-54}:{(74.2 - 50) / (80 - 50) * \R}) --
  ({-90}:{(44.6 - 30) / (48 - 30) * \R}) --
  ({-126}:{(88.6 - 80) / (90 - 80) * \R}) --
  ({-162}:{(65.1 - 50) / (69 - 50) * \R}) --
  ({162}:{(67.2 - 50) / (72 - 50) * \R}) --
  ({126}:{(76.1 - 55) / (81 - 55) * \R}) -- cycle;

\foreach \a/\v/\lo/\hi in {
  90/79.7/55/86, 54/82.8/55/90, 18/92.1/50/103,
  -18/63.0/50/66, -54/74.2/50/80, -90/44.6/30/48,
  -126/88.6/80/90, -162/65.1/50/69, 162/67.2/50/72,
  126/76.1/55/81} {
  \fill[colorA] (\a:{(\v - \lo) / (\hi - \lo) * \R}) circle (1.5pt);
  \pgfmathsetmacro{\anchA}{\a + 180}
  \node[font=\tiny, text=colorA!80!black, anchor=\anchA, inner sep=1.5pt]
    at (\a:{(\v - \lo) / (\hi - \lo) * \R}) {\v};
}

\draw[colorB, semithick, fill=colorB, fill opacity=0.15]
  ({ 90}:{(61.6 - 55) / (86 - 55) * \R}) --
  ({ 54}:{(61.4 - 55) / (90 - 55) * \R}) --
  ({ 18}:{(60.6 - 50) / (103 - 50) * \R}) --
  ({-18}:{(56.2 - 50) / (66 - 50) * \R}) --
  ({-54}:{(59.4 - 50) / (80 - 50) * \R}) --
  ({-90}:{(37.2 - 30) / (48 - 30) * \R}) --
  ({-126}:{(86.6 - 80) / (90 - 80) * \R}) --
  ({-162}:{(57.9 - 50) / (69 - 50) * \R}) --
  ({162}:{(57.1 - 50) / (72 - 50) * \R}) --
  ({126}:{(64.0 - 55) / (81 - 55) * \R}) -- cycle;

\foreach \a/\v/\lo/\hi in {
  90/61.6/55/86, 54/61.4/55/90, 18/60.6/50/103,
  -18/56.2/50/66, -54/59.4/50/80, -90/37.2/30/48,
  -126/86.6/80/90, -162/57.9/50/69, 162/57.1/50/72,
  126/64.0/55/81} {
  \fill[colorB] (\a:{(\v - \lo) / (\hi - \lo) * \R}) circle (1.5pt);
  \pgfmathsetmacro{\anchB}{\a + 180}
  \node[font=\tiny, text=colorB, anchor=\anchB, inner sep=1.5pt]
    at (\a:{(\v - \lo) / (\hi - \lo) * \R}) {\v};
}

\node[font=\footnotesize] at (0, {-\R - 0.8}) {%
  \textcolor{colorA}{\rule{6pt}{6pt}}~Direct (avg 73.3) \quad
  \textcolor{colorB}{\rule{6pt}{6pt}}~Two-Stage (avg 60.2)%
};

\end{tikzpicture}
  \caption{Direct vs.\ two-stage adaptation across 10 benchmarks. Each axis is independently scaled.}
  \label{fig:radar_dual_path}
  \vspace{-10pt}
\end{wrapfigure}

As discussed in Section~\ref{sec:dual_path}, we compare two AR-to-diffusion conversion strategies.
The \emph{two-stage} path starts from the AR LLM (Qwen2.5-Instruct-3B): it first fine-tunes on 300K text-only samples following the Fast-dLLM v2 recipe to obtain a diffusion LLM, then fine-tunes on multimodal data to produce a diffusion VLM.
The \emph{direct} path starts from the AR VLM (Qwen2.5-VL-3B) and converts it into a single multimodal fine-tuning stage.
The multimodal fine-tuning data, training recipe, and compute budget are identical across both paths.

Figure~\ref{fig:radar_dual_path} summarizes the comparison.
The direct path achieves a substantially higher average score (73.3 vs.\ 60.2), outperforming the two-stage path on all 10 benchmarks.
The gap is particularly large on knowledge- and reasoning-intensive tasks such as DocVQA (+31.5), ChartQA (+21.4), and AI2D (+18.1), where the pretrained multimodal alignment of the base VLM provides a strong advantage.

Notably, both paths consume \textbf{comparable training data} (2M training samples detailed in Appendix.\ref{app:training_details}) and \textbf{compute budget} (both trained single epoch to avoid memorization), yet the direct path outperforms by a wide margin (radar chart in Figure.\ref{fig:radar_dual_path}).
This is because the direct path inherits the multimodal alignment already acquired during VLM pretraining, making more efficient use of the same training budget, whereas the two-stage path starts from a text-only LLM and must rebuild this alignment from scratch.
We hypothesize that both strategies share a similar performance ceiling but differ primarily in training efficiency, motivating our choice of the direct path as the default recipe.

\subsection{Ablation Studies}
\label{sec:ablation}

We conduct ablation experiments under the direct conversion setting using the ShareGPT-4V dataset~\citep{chen2023sharegpt4v}.
Table~\ref{tab:ablation} isolates the contribution of each training recipe component by removing one at a time from the full configuration.

\paragraph{Causal context.}
Replacing the causal context attention with block-level bidirectional attention for all preceding context causes the most severe degradation, with average accuracy dropping by 22.5\% (from 57.3 to 44.4).
The damage is especially pronounced on reasoning-heavy benchmarks such as MMMU-Pro-V ($-$58.9\%) and SeedBench2+ ($-$39.5\%), confirming that causal context is essential for preserving the pretrained AR model's sequential reasoning capability and enabling effective AR-loss co-training.

\paragraph{Annealing block size.}
Training directly with the target block size $B_d=32$ without the curriculum schedule leads to a 4.4\% average drop.
The degradation is concentrated on long-form generation (MMMU-Pro-V $-$32.5\%), indicating that progressive exposure to larger blocks is critical for the model to learn stable denoising at large corruption spans.

\paragraph{Auto-truncation attention mask.}
Removing the auto-truncation mechanism causes a 3.7\% average drop, with a notable 14.4\% decline on MMMU.
Without truncation, the last block of each response extends into the next turn's prompt, leaking future information during training and degrading generation reliability at inference time.

\newcommand{\drop}[1]{\textcolor{red}{\scriptsize{(#1)}}}
\newcommand{\gain}[1]{\textcolor{green!60!black}{\scriptsize{(#1)}}}
\begin{table*}[t]
  \centering
  \caption{%
    Ablation study on training recipe.
    Each row removes one component from the full recipe.
    Parenthesized values show relative change: \textcolor{red}{red} for drops, \textcolor{green!60!black}{green} for gains.%
  }
  \label{tab:ablation}
  \vspace{4pt}
  \resizebox{\textwidth}{!}{%
    \begin{tabular}{l *{7}{c}}
      \toprule
      Setting
        & MMBench & MMMU & POPE & MMMU-Pro-V
        & RealWorldQA & SeedBench2+ & Avg \\
      \midrule
      Full recipe
        & 72.4   & 43.0   & 85.1   & 15.1
        & 61.1   & 66.9   & 57.3 \\
      \midrule
      w/o causal context
        & 58.5 \drop{-19.2\%}  & 29.9 \drop{-30.5\%}  & 71.1 \drop{-16.5\%}  & 6.2 \drop{-58.9\%}
        & 60.0 \drop{-1.8\%}   & 40.5 \drop{-39.5\%}  & 44.4 \drop{-22.5\%} \\
      w/o annealing block size
        & 68.6 \drop{-5.2\%}   & 43.4 \gain{+0.9\%}   & 81.4 \drop{-4.3\%}   & 10.2 \drop{-32.5\%}
        & 58.4 \drop{-4.4\%}   & 66.8 \drop{-0.1\%}   & 54.8 \drop{-4.4\%} \\
      w/o auto-truncation
        & 68.4 \drop{-5.5\%}   & 36.8 \drop{-14.4\%}  & 84.3 \drop{-0.9\%}   & 13.5 \drop{-10.6\%}
        & 61.0 \drop{-0.2\%}   & 67.1 \gain{+0.3\%}   & 55.2 \drop{-3.7\%} \\
      \bottomrule
    \end{tabular}%
  }%
\end{table*}

\subsection{Inference Acceleration}
\label{sec:acceleration}

\begin{wrapfigure}{r}{0.42\columnwidth}
  \vspace{-12pt}
  \centering
  \includegraphics[width=0.4\columnwidth]{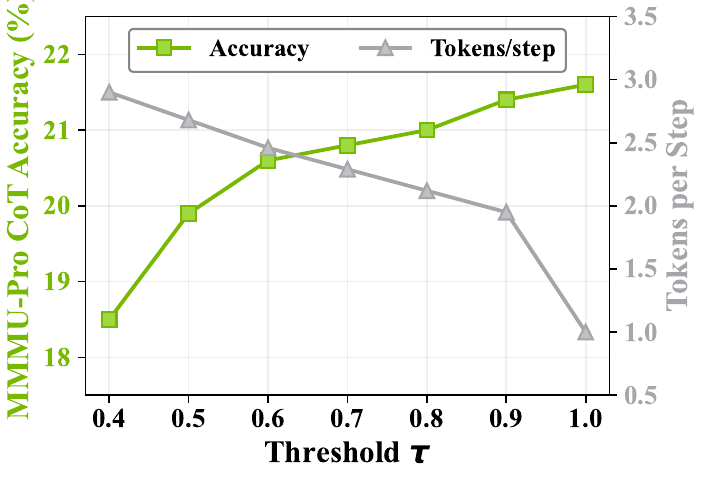}
  \caption{%
    Effect of threshold $\tau$ on MMMU-Pro CoT accuracy (left axis)
    and tokens per step (right axis).%
  }
  \label{fig:threshold_accuracy}
  \vspace{-12pt}
\end{wrapfigure}

We analyze three dimensions of inference acceleration: the confidence threshold $\tau$ that controls parallelism within MDM decoding, self-speculative decoding that further boosts throughput, and system-level integration with SGLang~\citep{zheng2024sglang} for production-grade serving.
Figure~\ref{fig:teaser}(c) and Table~\ref{tab:speed} summarize wall-clock speedup on MMMU-Pro-V, where long-form generation makes latency savings most visible.

\paragraph{Decoding threshold.}

The confidence threshold $\tau$ governs the speed--quality tradeoff in MDM decoding (Figure~\ref{fig:threshold_accuracy}).
We sweep $\tau$ from 0.4 to 1.0: at $\tau=1.0$, only one token is revealed per step (21.6 accuracy); relaxing to $\tau=0.9$ nearly doubles throughput to 1.95 tokens/step while maintaining 21.4 accuracy; pushing to $\tau=0.4$ maximizes parallelism at 2.90 tokens/step but accuracy drops to 18.5.
We adopt $\tau=0.9$ as the default for the best quality--speed balance.

\begin{figure*}[t]
  \centering
  \includegraphics[width=0.85\textwidth]{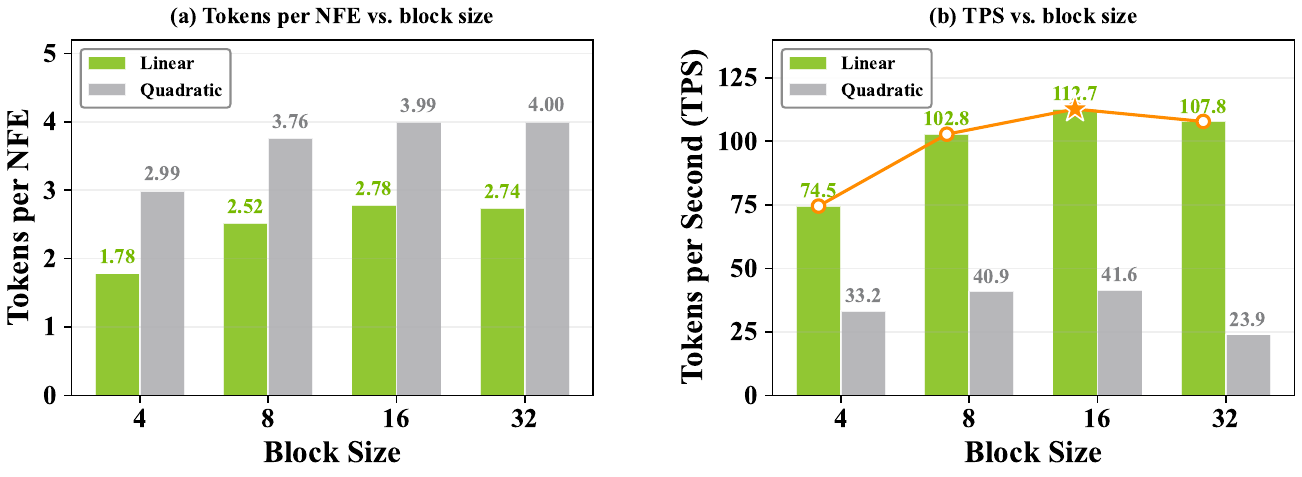}
  \caption{%
    Speculative decoding throughput across block sizes, comparing linear and quadratic variants.
    (a)~Tokens per NFE increases with block size for both variants.
    (b)~Wall-clock TPS peaks at block size 16 for both variants;
    the quadratic variant's $O(B^2)$ cost becomes dominant at larger block sizes.%
  }
  \vspace{-10pt}
  \label{fig:blocksize_comparison}
\end{figure*}

\paragraph{Speculative block decoding.}
Self-speculative decoding further improves both quality and speed: the speculative variant recovers accuracy to 24.6 (close to the AR baseline's 26.3) while achieving $1.98\times$ wall-clock TPS speedup (112.7 vs.\ 56.7 TPS).
Figure~\ref{fig:blocksize_comparison} compares the linear and quadratic variants across block sizes.
Tokens/NFE increases monotonically with block size for both variants, as larger blocks allow more tokens to be proposed per forward pass.
The linear variant peaks in wall-clock TPS at block size 16 (112.7 TPS) and slightly drops at 32, as per-step computation grows while the acceptance rate saturates.
The quadratic variant achieves higher Tokens/NFE thanks to its fused verify-and-propose design, but its TPS is lower at all block sizes: each step processes $\hat{B}\times(\hat{B}+1)$ tokens with a non-standard attention mask pattern that current kernels are not optimized for, making the theoretical NFE advantage hard to realize in wall-clock time.

\paragraph{SGLang integration.}
We integrate \modelname{} into the SGLang inference engine, extending its scheduler with block-diffusion attention masking: the draft step uses bidirectional attention within each block and the verify step uses causal attention, both sharing the same paged KV cache.
This allows \modelname{} to benefit from SGLang's optimized kernels, and CUDA graph.
On top of speculative decoding, SGLang serving further boosts throughput to 319.0 TPS. To reduce memory footprint and better leverage Tensor Core efficiency, we further enable SmoothQuant W8A8 (FP8) quantization, increasing throughput to 350.3 TPS ($6.18\times$ speedup, Figure~\ref{fig:teaser}(c) and Table~\ref{tab:speed}).

\section{Conclusion}

We presented \modelname{}, a block-diffusion vision-language model that converts a pretrained autoregressive VLM into diffusion VLM with KV-cache compatibility and self-speculative decoding.
Through a systematic comparison of two AR-to-diffusion conversion strategies, we showed that direct conversion from a fully pretrained AR VLM is both simpler and more effective than the two-stage approach, better preserving the multimodal capabilities acquired during pretraining.
We further proposed a comprehensive training recipe---including block-size annealing, causal context attention, auto-truncation masking, and vision-efficient concatenation---and validated the contribution of each component through ablation studies.
Experiments across 11 multimodal benchmarks demonstrated that \modelname{} matches its AR counterpart. Combined with SGLang integration and FP8 quantization, \modelname{} achieves over $6\times$ end-to-end inference speedup.

\bibliography{colm2026_conference}
\bibliographystyle{colm2026_conference}

\clearpage
\newpage
\appendix
\section{Experimental Setup}
\label{app:training_details}

\subsection{Training Data}

Our multimodal fine-tuning stage is conducted on a diverse mixture of instruction-tuning datasets, curated in reference to the NVILA~\citep{liu2024nvila} training mixture. This collection encompasses approximately 2 million training samples spanning general visual instruction tuning, document and chart understanding, as well as domain-specific scientific and geometric reasoning.
Specifically, the dataset mixture consists of the following components:
\begin{itemize}
    \item \textbf{General purpose and conversational data:} High-quality image-text instruction pairs sourced from ShareGPT4V~\citep{chen2023sharegpt4v} (including the GPT-4 generated 100K split and the broader SFT subset) and LLaVA-Instruct~\citep{liu2023visual}.
    \item \textbf{Chart and data visualization:} Specialized subsets from DVQA~\citep{kafle2018dvqa} and ChartQA~\citep{masry2022chartqa}.
    \item \textbf{Scientific and geometric reasoning:} Diagrams from AI2D~\citep{kembhavi2016ai2d} and geometric problem-solving questions from GeoQA~\citep{chen2021geoqa}.
    \item \textbf{Document understanding:} Document images and text-rich visual question answering samples from DocVQA~\citep{mathew2021docvqa} and synthetically generated document content via SynthDoG~\citep{kim2022ocr}.
\end{itemize}

\subsection{Training Configuration}

We follow the direct path (Section~\ref{sec:dual_path}), initializing from Qwen2.5-VL-3B~\citep{bai2025qwen25vl}.
Training is conducted on 64 NVIDIA H100 GPUs (8 nodes $\times$ 8 GPUs per node) using DeepSpeed ZeRO Stage-2 with BF16 mixed precision and gradient checkpointing enabled.
We train for 1 epoch with a cosine learning rate schedule, a peak learning rate of $5 \times 10^{-6}$, and a warmup ratio of 0.03.
The per-device batch size is 1 with gradient accumulation over 4 steps, yielding an effective global batch size of 256.
During training, the LLM backbone, vision encoder, and MLP projection layers are fine-tuned jointly.
We use block-size annealing (Section~\ref{sec:annealing_block}) with a target block size of $B_d=32$, complementary masking, causal context attention, and vision-efficient concatenation.
The diffusion and causal loss branches are weighted equally ($\alpha=\beta=0.5$).

\subsection{Benchmarks}

We evaluate on 11 VLM benchmarks spanning two categories.
\textit{Short-answer benchmarks} require brief, factual responses:
AI2D~\citep{kembhavi2016ai2d},
ChartQA~\citep{masry2022chartqa},
DocVQA~\citep{mathew2021docvqa},
GQA~\citep{hudson2019gqa},
MMBench~\citep{liu2024mmbench},
MMMU~\citep{yue2024mmmu},
POPE~\citep{li2023pope},
RealWorldQA~\citep{realworldqa2024},
SEEDBench2+~\citep{li2024seedbench2plus},
and TextVQA~\citep{singh2019textvqa}.
\textit{Long-answer benchmarks} require extended chain-of-thought reasoning:
MMMU-Pro-V~\citep{yue2025mmmupro}.

\subsection{Evaluation Protocol}

All benchmarks are evaluated using VLMEvalKit~\citep{duan2024vlmevalkit} under the same prompts and post-processing as the AR baseline.
Throughput (tokens per second, TPS) is measured on a single NVIDIA H100 GPU at batch size one, reflecting the single-request regime prevalent in physical AI deployments (e.g., robotics and autonomous driving) where AR decoding is memory-bandwidth-bound.
We also report Tokens/NFE (tokens per number of forward evaluations), which measures the average number of tokens decoded per forward pass.
Tokens/NFE on MMMU-Pro-V is computed only on samples with response length greater than 200 tokens, as this metric is less meaningful for short responses.
TPS speedup and Tokens/NFE are both reported relative to the AR baseline Qwen2.5-VL-3B.

\begin{table}[t]
  \centering
  \caption{%
    Inference acceleration on MMMU-Pro-V.
    Each row progressively adds one optimization on top of the previous. SpeedUp is relative to the AR baseline.%
  }
  \label{tab:speed}
    \begin{tabular}{@{} l c c c @{}}
      \toprule
      Setting
        & MMMU-Pro-V & TPS & SpeedUp \\
      \midrule
      AR baseline
        & 26.3   &  56.7  & 1.00$\times$ \\
      \midrule
      \modelname{} (MDM, $\tau{=}0.9$)
        & 21.4  & 82.2   & 1.45$\times$  \\
      \quad + Spec.\ decoding (linear)
        & 24.6   & 112.7   & 1.98$\times$  \\
      \quad\quad + SGLang serving
        & 24.1   & 319.0   & 5.63$\times$  \\
      \quad \quad \quad + SmoothQuant-W8A8 (FP8) & 23.8 & 350.3 & 6.18$\times$\\ 
      \bottomrule
    \end{tabular}
\end{table}

\section{Details of Speculative Decoding}
\label{sec:spec_details}

\begin{algorithm}[t]
    \caption{Linear Speculative Block-Causal Decoding}
    \label{alg:linear-spec}
    \begin{algorithmic}[1]
    \REQUIRE prompt $\mathbf{x}_{1:L}$, block size $B$, mask token \texttt{[M]}
    
    \STATE \textcolor{blue}{\textit{\% Prefill (causal attention)}}
    \STATE $\mathbf{h}, \mathrm{KV} \gets \mathrm{Forward}_{\text{causal}}(\mathbf{x}_{1:L})$
    \STATE $x_{L+1} \gets \arg\max\, \mathbf{h}_L$; \quad $n \gets L+1$
    
    \WHILE{not done}
        \STATE \textcolor{blue}{\textit{\% === Draft step: Bidirectional attention (Diffusion mode) ===}}
        \STATE \textcolor{blue}{\textit{\% Each token in the block attends to all other tokens in the block,}}
        \STATE \textcolor{blue}{\textit{\% and to all cached prefix tokens (via KV cache). No causal constraint.}}
        \STATE $\mathbf{d} \gets [x_n,\, \texttt{[M]},\, \dots,\, \texttt{[M]}]$ \hfill $\triangleright$ $B$ tokens: 1 real $+ (B{-}1)$ masks
        \STATE Attention mask $\mathbf{A}^{\text{draft}}$: $A_{ij} = 1 \;\;\forall\, i,j \in \{1,\dots,B\}$ \hfill $\triangleright$ Full bidirectional
        \STATE $\hat{\mathbf{h}} \gets \mathrm{Forward}(\mathbf{d},\, \mathrm{KV},\, \mathbf{A}^{\text{draft}},\, \textit{no cache update})$
        \STATE $\hat{x}_i \gets \arg\max\, \hat{\mathbf{h}}_{i-1}$ for $i=1,\dots,B$ \hfill $\triangleright$ Shift-by-one: $\hat{\mathbf{h}}_{i-1}$ predicts position $i$
        \STATE $d_i \gets \hat{x}_i$ for all mask positions $i$ \hfill $\triangleright$ Fill block with draft predictions
    
        \STATE \textcolor{blue}{\textit{\% === Verify step: Causal attention (AR mode) ===}}
        \STATE \textcolor{blue}{\textit{\% Standard left-to-right causal mask; each token only sees preceding tokens.}}
        \STATE Attention mask $\mathbf{A}^{\text{verify}}$: $A_{ij} = \mathbb{1}[i \ge j]$ \hfill $\triangleright$ Causal (lower triangular)
        \STATE $\mathbf{h}^{\text{v}}, \mathrm{KV} \gets \mathrm{Forward}(\mathbf{d},\, \mathrm{KV},\, \mathbf{A}^{\text{verify}},\, \textit{update cache})$
        \STATE $a_i \gets \arg\max\, \mathbf{h}^{\text{v}}_i$ for $i=0,\dots,B{-}1$ \hfill $\triangleright$ AR predictions
    
        \STATE \textcolor{blue}{\textit{\% === Accept: left-to-right comparison ===}}
        \STATE $k \gets \min\{j \in [0, B{-}2] : a_j \neq d_{j+1}\}$;\; $k \gets k + 1$ \hfill $\triangleright$ First mismatch; accept $k$ tokens
        \STATE Append $a_{0:k-1}$ to sequence; $n \gets n + k$
        \STATE $\mathrm{KV} \gets \mathrm{Crop}(\mathrm{KV},\, n - 1)$ \hfill $\triangleright$ Discard rejected tokens from cache
    \ENDWHILE
    
    \RETURN generated sequence
    
    \end{algorithmic}
\end{algorithm}

\begin{algorithm}[t]
    \caption{Quadratic Speculative Block-Causal Decoding}
    \label{alg:quadratic-spec}
    \begin{algorithmic}[1]
    \REQUIRE prompt $\mathbf{x}_{1:L}$, block size $B$, mask token \texttt{[M]}
    
    \STATE \textcolor{blue}{\textit{\% === Step 1: Draft-only forward ===}}
    \STATE \textcolor{blue}{\textit{\% Prompt tokens use causal attention; mask tokens attend to everything (bidirectional).}}
    \STATE $\mathbf{z} \gets [\mathbf{x}_{1:L},\, \texttt{[M]}^B]$
    \STATE Attention mask: $A_{ij} = \begin{cases} \mathbb{1}[i \ge j] & \text{if } i < L \text{ (prompt: causal)} \\ 1 & \text{if } i \ge L \text{ (mask block: attend to all)} \end{cases}$
    \STATE $\mathbf{h}, \mathrm{KV} \gets \mathrm{Forward}(\mathbf{z},\, \mathbf{A})$
    \STATE $d^{(1)}_i \gets \arg\max\, \mathbf{h}_{L-1+i}$ for $i=0,\dots,B{-}1$ \hfill $\triangleright$ First block draft
    \STATE $\mathrm{KV} \gets \mathrm{Crop}(\mathrm{KV},\, L)$ \hfill $\triangleright$ Only keep prompt cache
    
    \STATE \textcolor{blue}{\textit{\% === Step 2: Quadratic verify + propose loop ===}}
    \STATE $s \gets 0$ \hfill $\triangleright$ Total accepted tokens
    \WHILE{$s < T$}
        \STATE \textcolor{blue}{\textit{\% Construct quadratic input: expand $B$ draft tokens into $B$ groups of $(B{+}1)$}}
        \STATE Given draft block $\mathbf{d} = [d_0, d_1, \dots, d_{B-1}]$
        \STATE Build input: $\mathbf{q} = [\underbrace{d_0, \texttt{[M]}^{B}}_{\text{group 0}},\; \underbrace{d_1, \texttt{[M]}^{B}}_{\text{group 1}},\; \dots,\; \underbrace{d_{B-1}, \texttt{[M]}^{B}}_{\text{group } B{-}1}]$
        \STATE \hfill $\triangleright$ Total length: $B \times (B+1)$
    
        \STATE \textcolor{blue}{\textit{\% Position IDs: group $i$ gets positions $[s{+}i,\; s{+}i{+}1,\; \dots,\; s{+}i{+}B]$}}
        \STATE \textcolor{blue}{\textit{\% This simulates ``if accepted up to $d_i$, generate next $B$ tokens''}}
    
        \STATE \textcolor{blue}{\textit{\% Attention mask structure (within the $B(B{+}1)$ query tokens):}}
        \STATE \textcolor{blue}{\textit{\% (a) All tokens attend to entire KV cache (prefix)}}
        \STATE \textcolor{blue}{\textit{\% (b) First token of each group ($d_i$) attends causally to first tokens of groups $\le i$}}
        \STATE \textcolor{blue}{\textit{\% (c) Non-first tokens in group $i$ attend to all tokens in the same group (bidirectional)}}
        \STATE \textcolor{blue}{\textit{\% (d) No cross-group attention for non-first tokens}}
    
        \STATE $\mathbf{H}, \mathrm{KV'} \gets \mathrm{Forward}(\mathbf{q},\, \mathrm{KV},\, \mathbf{A}^{\text{quad}})$
        \STATE Reshape: $\mathbf{H} \in \mathbb{R}^{B \times (B+1) \times |V|}$
        \STATE $P_{i,j} \gets \arg\max\, \mathbf{H}_{i,j}$ \hfill $\triangleright$ Group $i$, position $j$
    
        \STATE \textcolor{blue}{\textit{\% Verify: $P_{i,0}$ is the AR prediction for position $s{+}i{+}1$}}
        \STATE $k \gets 1$
        \STATE $\mathbf{d}^{\text{next}} \gets P_{0,\, 1:B}$ \hfill $\triangleright$ Default: next-block proposal from group 0
        \WHILE{$k < B$ \AND $P_{k-1,\, 0} = d_k$}
            \STATE $\mathbf{d}^{\text{next}} \gets P_{k,\, 1:B}$ \hfill $\triangleright$ Update proposal from latest accepted group
            \STATE $k \gets k + 1$
        \ENDWHILE
    
        \STATE Accept $d_{0:k-1}$;\; $s \gets s + k$
        \STATE $\mathbf{d} \gets \mathbf{d}^{\text{next}}$ \hfill $\triangleright$ Next block's draft
        \STATE $\mathrm{KV} \gets$ extract draft-token-only KVs from $\mathrm{KV'}$ \hfill $\triangleright$ Keep every $(B{+}1)$-th entry
    \ENDWHILE
    
    \RETURN $\mathbf{x}_{1:L+s}$
    \end{algorithmic}
    \end{algorithm}

We describe two self-speculative decoding strategies for block-causal diffusion VLMs.
Both exploit the fact that our model supports two attention modes within a single set of weights: a \emph{bidirectional} (diffusion) mode for parallel drafting and a \emph{causal} (AR) mode for verification.
The diffusion mode produces a draft of an entire block in one pass, and the causal mode verifies its consistency with the AR distribution---realizing the draft-then-verify paradigm within a single model.

\paragraph{Linear speculative decoding}
(Algorithm~\ref{alg:linear-spec}) processes each block with exactly two forward passes.
In the \emph{draft} step, $B{-}1$ mask tokens are appended after the last accepted token to form a block of size $B$, and a forward pass with bidirectional attention predicts all masked positions simultaneously.
In the \emph{verify} step, the filled block is re-evaluated with causal attention, yielding standard AR predictions.
The draft and AR predictions are compared left to right: consecutive matches are accepted, and the first mismatch is replaced by the AR prediction.
The KV cache is cropped to the accepted length before the next iteration.
This requires 2~NFEs per block and can accept up to $B$ tokens, giving a theoretical speedup of up to $B/2\times$ over AR decoding.

\paragraph{Quadratic speculative decoding}
(Algorithm~\ref{alg:quadratic-spec}) fuses verification and proposal into a single forward pass.
The $B$ draft tokens $[d_0, d_1, \dots, d_{B-1}]$ are expanded into $B$ groups of $B{+}1$ tokens: group $i$ is $[d_i, \texttt{[M]}^{B}]$, yielding a total input of $B(B{+}1)$ tokens.
The attention mask ensures that (a)~the first token of each group attends causally to first tokens of preceding groups (AR verification), (b)~the $B$ mask tokens within each group attend bidirectionally to all tokens in the same group (parallel proposal), and (c)~all tokens attend to the prefix via the KV cache.
Position IDs for group $i$ cover $[s{+}i,\dots,s{+}i{+}B]$, simulating an AR rollout from each acceptance point.
After the forward pass, the output is reshaped to $B\times(B{+}1)\times|V|$: entry $(i,0)$ verifies position $i{+}1$, while entries $(i,1{:}B)$ provide the next-block proposal conditioned on accepting through position $i$.
This requires only 1~NFE per block at the cost of $O(B^2)$ input tokens per step.

\paragraph{Trade-offs.}
The linear variant scales linearly with $B$ and suits larger block sizes; the quadratic variant halves the number of forward passes but its $O(B^2)$ cost is more attractive for moderate $B$.
Both variants produce the same token sequence as standard block-causal generation---the choice is purely a latency optimization depending on hardware and block size.
\section{Case Study}
\label{app:case_study}

We present qualitative examples to illustrate how \modelname{} (speculative decoding) compares with the AR baseline in terms of both response quality and decoding efficiency.

\begin{figure}[t]
\centering
\includegraphics[width=\textwidth]{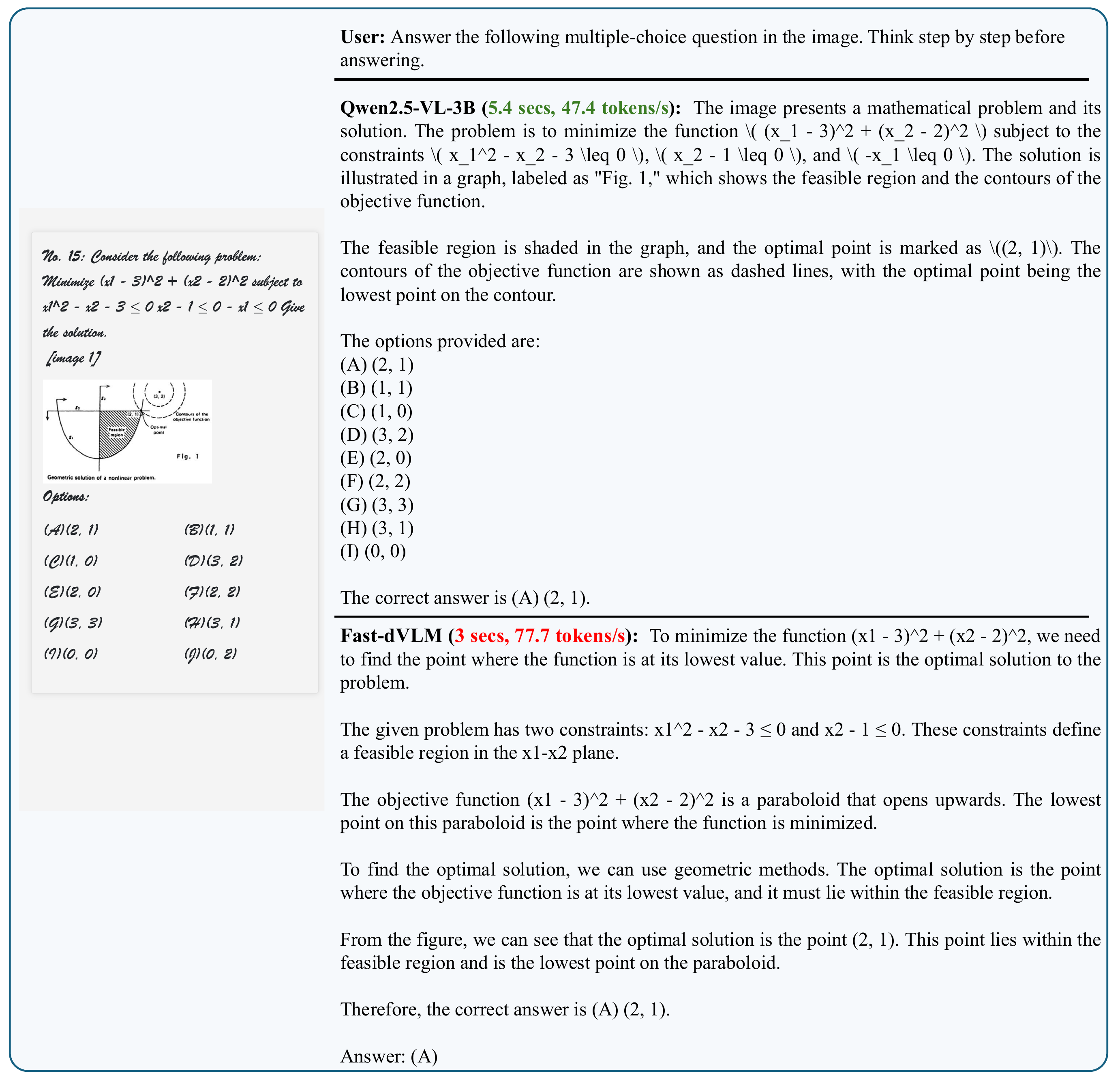}
\caption{%
  Qualitative comparison on an MMMU-Pro-V math reasoning problem.
  Both the AR baseline (Qwen2.5-VL-3B) and \modelname{} (speculative decoding) arrive at the correct answer, while \modelname{} achieves $1.6\times$ faster decoding speed (77.7 vs.\ 47.4 tokens/s).
}
\label{fig:case_study}
\end{figure}

\paragraph{Math reasoning.}
Figure~\ref{fig:case_study} shows a constrained optimization problem from MMMU-Pro-V.
Both the AR baseline (Qwen2.5-VL-3B) and \modelname{} correctly identify the optimal point $(2, 1)$ and produce coherent step-by-step reasoning.
Notably, the AR baseline renders its response in raw \LaTeX{} markup (e.g., \texttt{\textbackslash(} and \texttt{\textbackslash leq}), whereas \modelname{} outputs clean, human-readable mathematical notation.
Despite generating a comparable amount of reasoning, \modelname{} completes the response in 3 seconds at 77.7 tokens/s, a $1.6\times$ speedup over the baseline's 5.4 seconds at 47.4 tokens/s.

\begin{figure}[t]
\centering
\includegraphics[width=\textwidth]{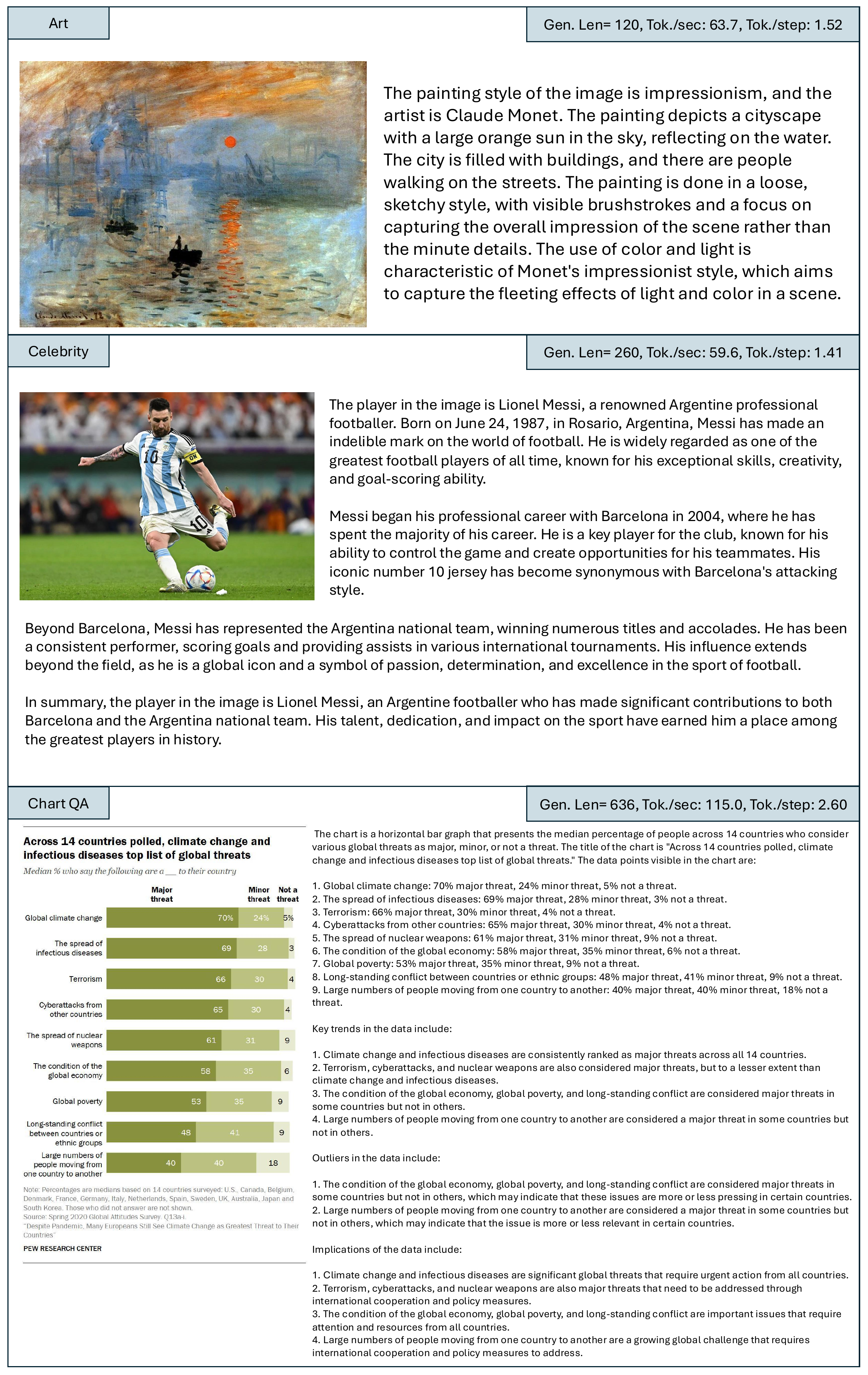}
\caption{%
  Additional qualitative examples spanning art style recognition, celebrity identification, and chart QA. \modelname{} produces accurate responses while maintaining high throughput.
}
\label{fig:more_case_study}
\end{figure}

\paragraph{Diverse task categories.}
Figure~\ref{fig:more_case_study} provides additional examples spanning art style recognition, celebrity identification, and chart question answering.
Across all three categories, \modelname{} generates detailed, accurate, and fluent responses:
it correctly identifies the impressionist style and attributes the painting to Claude Monet;
it recognizes Lionel Messi and provides extensive biographical context;
and it comprehensively reads the chart data, identifies key trends, and discusses implications.
Importantly, the decoding throughput remains high across different response lengths, ranging from 63.7 tokens/s for a short 120-token art description to 115.0 tokens/s for a long 636-token chart analysis, with Tokens/NFE ratios consistently above 1.5.
These examples confirm that \modelname{} preserves the generation quality of the AR baseline while delivering substantial inference speedup across diverse visual understanding tasks.

\begin{figure}[t]
\centering
\includegraphics[width=\textwidth]{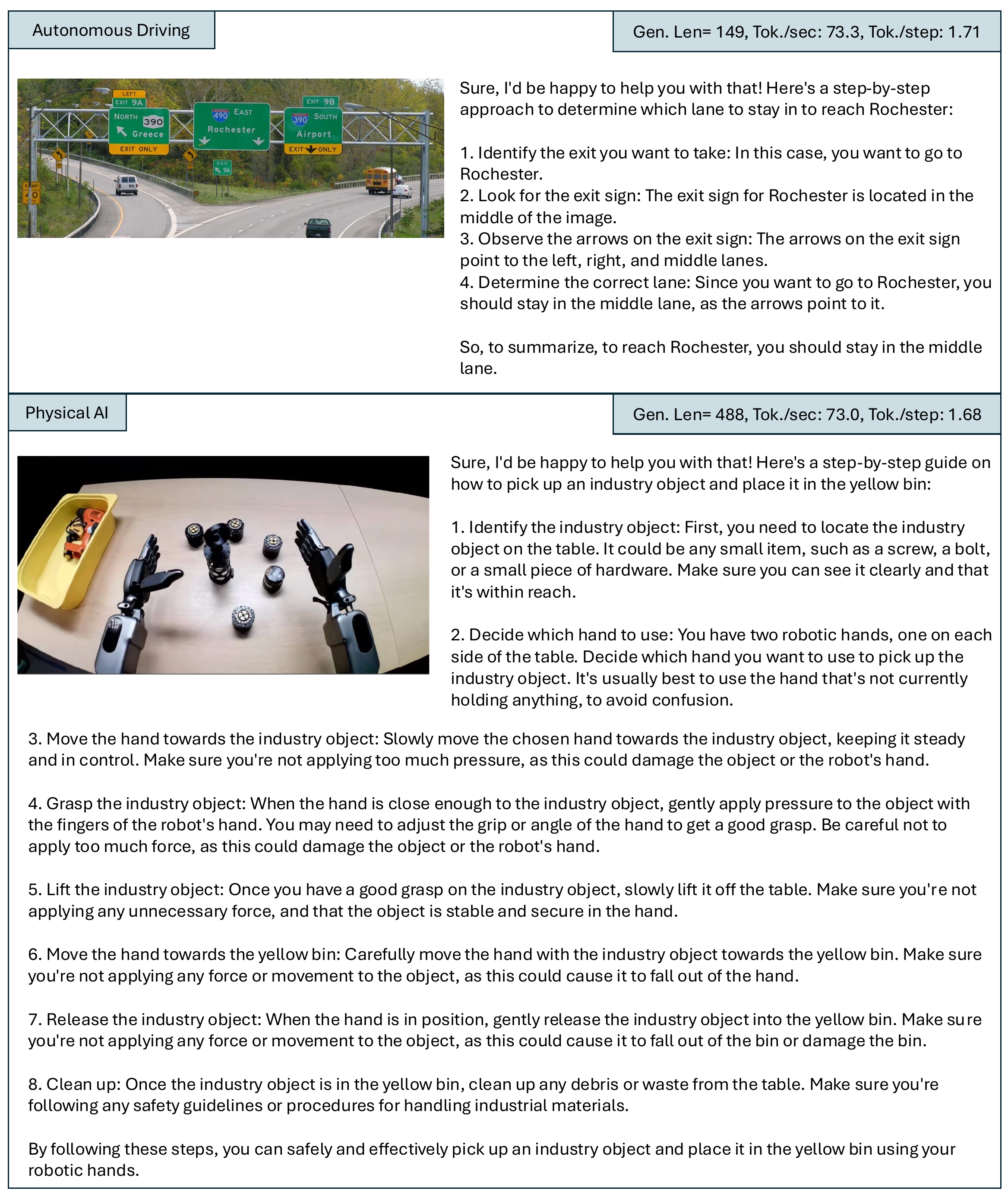}
\caption{%
  Qualitative examples of \modelname{} (speculative decoding) on embodied and physical AI tasks: autonomous driving scene understanding and robotic manipulation instruction.
  For each example, we report the generated response length (Gen.\ Len), decoding throughput (Tok./sec), and tokens per decoding step (Tok./step).
  \modelname{} produces detailed, step-by-step reasoning for both tasks while maintaining high throughput ($\sim$73 tokens/s) and a Tok./step ratio above 1.68.
}
\label{fig:physical_ai_case_study}
\end{figure}

\paragraph{Physical AI.}
Figure~\ref{fig:physical_ai_case_study} demonstrates \modelname{} on two embodied AI scenarios.
In the autonomous driving example, the model correctly reads highway signage and reasons about lane selection to reach Rochester, producing a concise 149-token response at 73.3 tokens/s.
In the robotic manipulation example, the model generates a detailed 488-token, 8-step guide for picking up an industrial object and placing it into a bin, maintaining 73.0 tokens/s throughput.
Both examples achieve a Tok./step ratio above 1.68, confirming that the block-diffusion speedup generalizes to long-form embodied reasoning tasks.

\section{Extended Related Work}
\label{app:related_work}

\subsection{Diffusion LLMs}

Discrete diffusion language models have evolved from continuous-latent formulations~\citep{lovelace2023latent,strudel2022self} to discrete masked diffusion objectives~\citep{austin2021structured,lou2023discrete,ou2024your,gulrajani2023likelihood,sahoo2024simple}, with recent models such as LLaDA~\citep{nie2025large} and Dream~\citep{ye2025dream} scaling to 7--8B parameters and matching LLaMA-3~\citep{grattafiori2024llama} across reasoning benchmarks; d1~\citep{zhao2025d1} further shows that dLLMs benefit from reinforcement learning.
A key remaining challenge is inference latency due to the incompatibility of bidirectional attention with KV caching.
Block-level generation addresses this by autoregressively producing blocks whose internal tokens are denoised in parallel: Block Diffusion~\citep{arriola2025block} enables KV caching and flexible-length generation; Fast-dLLM~\citep{wu2025fast} and Fast-dLLM v2~\citep{wu2025fastv2} introduce block-wise caching and efficient AR-to-diffusion adaptation recipes; and Set Block Decoding~\citep{gat2025sbd} combines next-token and masked-token prediction for 3--5$\times$ fewer forward passes.
With dLLMs now competitive on text-only tasks, extending this paradigm to vision-language settings is a natural next step.

\subsection{Diffusion VLMs}

Several concurrent works extend discrete diffusion to vision-language settings.
LLaDA-V~\citep{you2025llada} projects visual features into a masked diffusion LLM and matches strong AR baselines such as LLaMA3-V~\citep{grattafiori2024llama} and Qwen2-VL~\citep{wang2024qwen2};
LaViDa~\citep{li2025lavida} introduces complementary masking and prefix KV caching to accelerate multimodal diffusion decoding;
MMaDA~\citep{yang2025mmada} unifies multimodal reasoning and generation with a modality-agnostic architecture and RL-based post-training;
and Dimple~\citep{yu2025dimple} adopts a hybrid AR-then-diffusion paradigm with confident decoding.
However, all these models rely on full-sequence diffusion without block structure, precluding incremental KV caching of resolved response blocks, and none address turn-aware attention masking for multi-turn dialogue.

\subsection{Speculative Decoding for Diffusion LLMs}
Speculative decoding~\citep{leviathan2023fast,chen2023accelerating} accelerates inference by drafting multiple tokens and verifying them in parallel.
Recent work adapts this to dLLMs by exploiting their native multi-token prediction: SSD~\citep{gao2025self} uses the dLLM itself as both drafter and verifier; Spiffy~\citep{agrawal2025spiffy} proposes auto-speculation via directed draft graphs; BlockSpec~\citep{pan2025blockspec} introduces block-level speculation with dynamic token exploration; DFlash~\citep{chen2026dflash} and FailFast~\citep{pan2025failfast} integrate lightweight diffusion drafters; and DiffuSpec~\citep{li2025diffuspec} shows that a pretrained dLLM can serve as a training-free drafter for AR verifiers.
Our work integrates block-level self-speculation into a multimodal diffusion VLM for the first time.

\end{document}